**Video-to-Text Pedestrian Monitoring (VTPM): Leveraging Computer Vision and Large Language Models for Privacy-Preserve Pedestrian Activity Monitoring at Intersections**

**Ahmed S. Abdelrahman**
Corresponding Author
PhD Candidate
Smart and Safe Transportation Lab
Department of Civil, Environmental and Construction Engineering,
University of Central Florida, Orlando, FL, USA, 32816
Email: ahmed.abdelrahman@ucf.edu

**Mohamed Abdel-Aty**
Pegasus Professor and Trustee Chair
Smart and Safe Transportation Lab
Department of Civil, Environmental and Construction Engineering,
University of Central Florida, Orlando, FL, USA, 32816
Email: m.aty@ucf.edu

**Dongdong Wang, Ph.D.**
Smart and Safe Transportation Lab
Department of Civil, Environmental and Construction Engineering,
University of Central Florida, Orlando, FL, USA, 32816
Email: dongdong.wang@ucf.edu

**Abstract**

Computer vision has advanced research methodologies, enhancing system services across various fields. It is a core component in traffic monitoring systems for improving road safety; however, these monitoring systems don't preserve the privacy of pedestrians who appear in the videos, potentially revealing their identities. Addressing this issue, our paper introduces Video-to-Text Pedestrian Monitoring (VTPM), which monitors pedestrian movements at intersections and generates real-time textual reports, including traffic signal and weather information. VTPM uses computer vision models for pedestrian detection and tracking, achieving a latency of 0.05 seconds per video frame. Additionally, it detects crossing violations with 90.2% accuracy by incorporating traffic signal data. The proposed framework is equipped with Phi-3 mini-4k to generate real-time textual reports of pedestrian activity while stating safety concerns like crossing violations, conflicts, and the impact of weather on their behavior with latency of 0.33 seconds. To enhance comprehensive analysis of the generated textual reports, Phi-3 medium is fine-tuned for historical analysis of these generated textual reports. This fine-tuning enables more reliable analysis about the pedestrian safety at intersections, effectively detecting patterns and safety critical events. The proposed VTPM offers a more efficient alternative to video footage by using textual reports reducing memory usage, saving up to 253 million percent, eliminating privacy issues, and enabling comprehensive interactive historical analysis.

# 1. Introduction

Intelligent systems powered by Artificial Intelligence (AI) are revolutionizing the field of traffic safety and transportation, and their impact on pedestrian safety is particularly significant (Gangwani and Gangwani, 2021). Pedestrian safety holds a critical place in the broader context of traffic safety and transportation planning (El Hamdani et al., 2020). As cities expand and transportation networks become more complex, the importance of safeguarding pedestrians, often the most vulnerable road users (WHO, 2013), cannot be overstated. Ensuring pedestrian safety is not just a matter of reducing fatalities and injuries but also a foundational element in the pursuit of sustainable urban living and mobility (Aceves-González et al., 2020). Globally, pedestrian safety is a pressing concern, given the alarming statistics of accidents, where in 2020, the United States experienced 6,516 pedestrian deaths due to traffic accidents, marking a 3.9% rise from the 6,272 fatalities recorded in 2019. Additionally, approximately 54,769 pedestrians sustained injuries, according to the National Highway Traffic Safety Administration (NHTSA, 2022). Pedestrian fatalities represent 17% of all traffic fatalities, with an increase of 53% in 2018 compared to 2009 (GHSA, 2020). Pedestrians constitute a significant percentage of traffic fatalities worldwide (Zegeer and Bushell, 2012), highlighting the urgent need for comprehensive safety measures. This situation demands a multifaceted approach, integrating engineering, education, and enforcement to create safer urban environments.

In the pursuit of enhancing pedestrian safety and creating more efficient transportation systems, the integration of intelligent systems powered by AI is emerging as a game-changer. As cities continue to grow and urbanize, the complexities of managing pedestrian traffic and ensuring their safety on our roads have become increasingly challenging (Cervero et al., 2017). Fortunately, AI-driven technologies offer promising solutions to address these challenges, ushering in a new era of intelligent transportation systems (De Souza et al., 2018; El Hamdani et al., 2020). Many systems have been addressing pedestrian safety by monitoring their movements and actions at roads especially at intersections, where it is the road junction that captures all road users (Eom and Kim, 2020). These systems monitor pedestrian activity and estimate any safety related concerns, since they are vulnerable to any interaction with other motorized vehicles (Miranda-Moreno et al., 2011). Among various sensors integrated at intersections to monitor pedestrian activity, cameras is the most dominant sensor for this purpose (Mirhashemi et al., 2022). Surveillance cameras are installed at intersections to monitor pedestrians and their activity at intersections (Jiao and Fei, 2023; Yang et al., 2022; Zhang et al., 2020). However, the video footage often reveals the identities of the observed pedestrians, which raises significant privacy concerns (Slobogin and Brayne, 2023). Although these systems utilize video footage to extract key information such as pedestrian count, volume, and safety analysis, the failure to preserve individual identities is a major privacy issue in computer-vision tasks involving pedestrians (Ghaziamin et al., 2024). The ability to identify individuals from video footage not only violates personal privacy but also exposes individuals to potential risks, such as data breaches, and misuse of personal information (Vakaliuk et al., 2023). Therefore, effective ways for preserving pedestrians' privacy are needed to mitigate such risks without reducing the functionality of pedestrian monitoring systems.

Recently, Large language models (LLMs) have emerged as transformative tools across a wide range of domains, thanks to their ability to process and generate human-like text with remarkable fluency. These models, such as OpenAI's GPT-4, Google's Gemini, and Meta's Llama, are trained on vast datasets, enabling them to understand and produce language in a way that closely mimics human communication. Their applications extend beyond traditional natural language processing tasks like translation and summarization to more complex areas such as automated data curation, document summarization and analysis, and medical diagnostics (Acharya et al., 2023; Sedaghat, 2023; Yang et al., 2023). Their "large" designation is justified by their extensive training on massive amounts of text data and the integration of billions or trillions of tokens from different domains and topics on the internet, enabling them to perform complex language tasks (e.g., question-answering, summarization, text completion, emotion analysis, and problem-solving) with exceptional accuracy. Moreover, LLMs can be fine-tuned for certain tasks or datasets by training them on smaller, more targeted datasets from specific domains or topics (Chen and Mueller, 2024). This fine-tuning process highlights the flexibility and potential of LLMs for specialized applications. As a result, LLMs offer a more user-friendly interface and improved reasoning capabilities, benefiting from large-scale training that enhances their knowledge base and enables them to tackle increasingly complex linguistic tasks.

In this study, we propose Video-to-Text Pedestrian Monitoring (VTPM), a novel approach that integrates the strengths of computer vision for capturing visual data with the advanced textual analysis capabilities of LLMs. By employing camera-based computer vision to monitor pedestrian activity, VTPM shifts the analysis from video to text, harnessing the sophisticated reasoning power of LLMs to provide a comprehensive and privacy-preserving analysis of the pedestrian activity at intersections by extracting the required information from the video and transform into text that does not reveal the identity of any road user. Additionally, this approach allows for years of historical textual data to be stored and analyzed while dramatically reducing memory requirements to enable much rich analysis at locations of interest. The remainder of the paper is structured as follows: Section 2 reviews relevant studies on privacy-preserving pedestrian monitoring, Section 3 outlines the methodology of the VTPM, Section 4 discusses the results, and Section 5 offers the conclusion and future recommendations.

## 2. Related Work

Visual privacy-preserving methods hide some of the original information in line with regulatory data protection laws and individual privacy needs. Image filtering techniques such as blurring (Frome et al., 2009) and pixelation (Boyle et al., 2000) are widely used to improve privacy. The person can be easily identified using visual ques, such as clothing information even if face is completely covered with black box. Visual data should obfuscate all private information to provide complete privacy. Zheng et al. (Zhang et al., 2010) introduces a system that protects privacy in video surveillance by encrypting pedestrians' faces/heads. Utilizing a two-state Markov chain for tracking and elliptical head contour detection, the system obscures faces with a unique encryption code. This allows authorized personnel to selectively reveal identities while maintaining privacy for others. Sun et al. (2018) presents a technique to protect privacy in social media photos by generating realistic head inpainting. This method uses facial landmarks to inpaint new heads that blend seamlessly with the background, outperforming traditional blurring and blacking out methods in both natural appearance and identity obfuscation.

Although face blurring techniques are useful and can blurry the faces, blurring only the face might not be insufficient for privacy preservation because they fail to completely obfuscate identity information (Agarwal et al., 2021). Even if advanced techniques are used for face blurring still the pedestrian can be identified from his personal appearance (Brkic et al., 2017). As such others investigated other possibilities by mask the human body with an avatar representation. Kunchala et al. (2023) introduced a framework that uses 3D avatar reconstruction and digital in-painting to anonymize pedestrians in video data. This method preserves pose, shape, and background context while enhancing privacy compared to traditional blurring and pixelation. De Coninck et al. (2024) proposes a novel approach to privacy-preserving visual analysis using Generative Adversarial Privacy (GAP) techniques. The method obfuscates video data without requiring sensitive labels, ensuring privacy while enabling accurate visual analysis. Ohno et al. (2024) introduces a privacy-preserving pedestrian tracking system that employs multiple cameras and 3D point cloud features. By using path image inpainting, the system ensures pedestrian anonymity while maintaining accurate tracking capabilities. However, these methods are computationally expensive and typically are limited to generate avatar for single person or the allocation of the mask might not be so accurate revealing the identity of the pedestrians even for several video frames.

Other studies investigated processing videos on edge devices. Yuan et al. (2019) propose a pedestrian detection system that leverages edge computing and differential privacy to protect sensitive information in smart cities. The system processes data collected by IoT devices on edge servers instead of cloud servers, reducing latency and the risk of data breaches during transmission. By applying differential privacy using the Gaussian Mechanism, the system ensures pedestrian data privacy while maintaining high detection accuracy. Ghaziamin et al. (2024) employs a privacy-preserving methodology by processing data locally on solar-powered Nvidia Jetson edge devices, thus avoiding cloud transmission. It uses overhead fisheye cameras to capture images without identifying individual faces, leveraging computer vision models (YOLO-V4, DetectNet-V2, Faster-RCNN) optimized for edge deployment.

These studies underscore substantial research efforts to protect pedestrian privacy and secure their identities. However, methods such as avatar-based masking techniques are computationally demanding, and blurring techniques may not sufficiently guarantee complete preservation of identifiable features. Moreover, performing all the necessary computationally intensive processes directly on edge devices is impractical due to their limited computational power and memory resources (Lee et al., 2019). Privacy preservation in computer vision tasks is challenging, as it necessitates ensuring full protection of identifiable information while maintaining the functionality and effectiveness of the developed systems.

## 3. Methodology

In this paper, Video-to-Text Pedestrian Monitoring (VTPM) is introduced to address the key issues related to camera surveillance for monitoring pedestrian activity, particularly at intersections. These issues include preserving the privacy of all road users, reducing the economic burden of storing large amounts of video data, and facilitating extensive historical analysis. The proposed methodology ensures the preservation of road users' privacy, minimizes storage requirements, thereby promoting sustainability for future smart cities, generates report in real-time, and allows for comprehensive historical analysis for deeper insights at any road segment. VTPM allows for the full functionality of computer vision tasks and traffic safety measures without overly complicating the de-identification of pedestrians in videos. Key features extracted from the analyzed videos are converted into textual reports using a large language model, facilitating historical analysis that can be extended to years of saved textual data.

### 3.1 VTPM Pipeline

VTPM pipeline consists of three primary components, as represented in Fig. 1. The first component is pedestrian activity monitoring, which utilizes state-of-the-art deep learning and computer vision techniques and models to observe pedestrian activity and the interactions between pedestrians and turning vehicles. These interactions with vehicles that are turning, particularly to the right, are frequently the cause of most pedestrian-vehicle near-miss incidents, potentially leading to significant accidents (Buch and Jensen, 2017; Chen et al., 2022). Such incidents often occur when distracted drivers fail to notice pedestrians crossing during their right turn, despite pedestrians proceeding based on the controlled system's go-ahead during their designated crossing phase. This is the critical moment when most near misses occur between turning vehicles and pedestrians. It is important to note that this problem of hazardous interactions also extends to autonomous vehicles (AVs), where the engagements between pedestrians and right-turning AVs pose the greatest danger (Alozi and Hussein, 2023), primarily due to either pedestrians violating crossing rules or the limitations of the AVs' perception systems in detecting and predicting pedestrian crossings. To capture the pedestrian activity at the waiting area at the intersection and their interactions with right-turning vehicles, a camera is placed on the mast arm of the traffic pole to monitor the ongoing events.

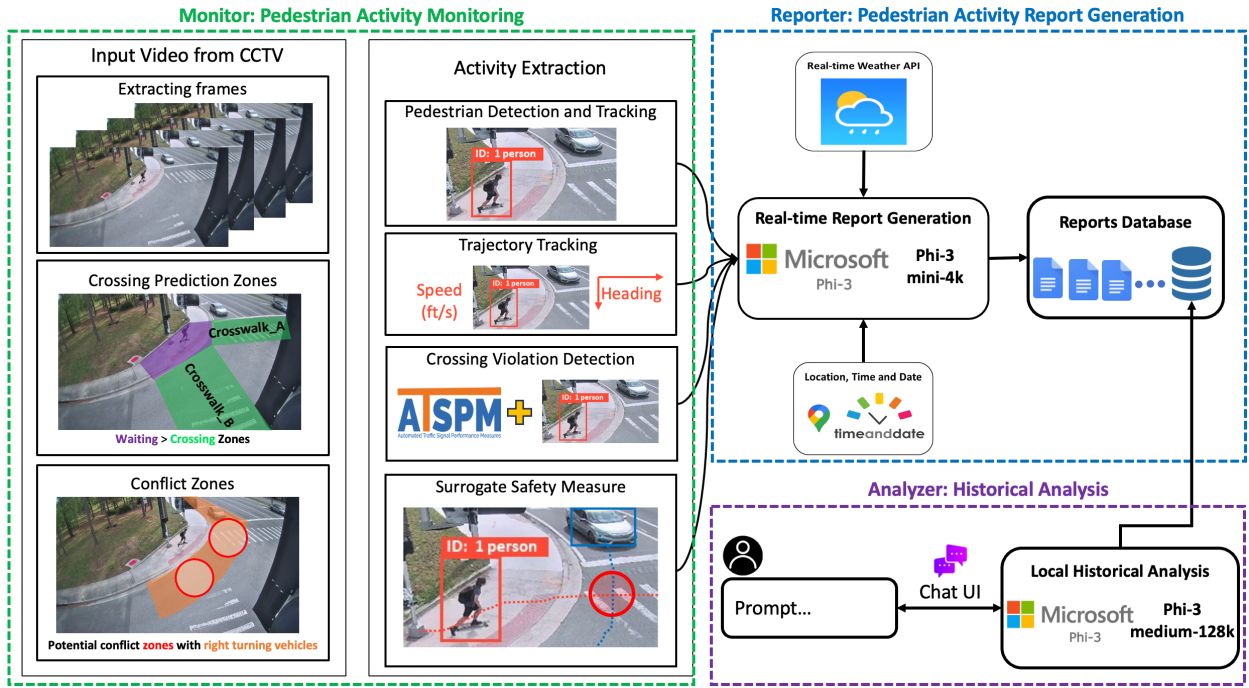

**Fig. 1. Video-to-text pipeline incorporating the monitor, the reporter, and the analyzer**

All raw data extracted from the pedestrian activity monitoring component will be forwarded to an LLM along with weather, date, time, and location information at the intersection to generate a textual report about pedestrian activity. Critical weather factors that influence pedestrian and driver behavior (Pińskwar et al., 2024; Zhai et al., 2019), including temperature, humidity, and rain intensity (classified as no rain, light rain, moderate rain, and heavy rain), will be imported via the Application Programming Interface (API) from OpenWeatherMap. To enable fast report generation for real-time purposes, a fast LLM is required. For this purpose, Microsoft's phi-3-mini, which has 3.8 billion parameters and a context window of 4096 tokens, will be utilized to support real-time textual report generation (Abdin et al., 2024).

The third component facilitates local historical analysis, which can be conducted once a substantial number of these reports have been gathered. This analysis leverages the collected data to derive valuable insights, identifying anomalies, patterns, safety issues, and other relevant information. Analysts interact dynamically with the LLM through a chat-based user interface, enabling a comprehensive and intuitive exploration of the data. Utilizing Microsoft's phi-3-medium model with 14 billion parameters for historical analysis purposes (Abdin et al., 2024), this component allows for in-depth investigations and continuous monitoring of pedestrian and vehicle interactions over time. This approach enhances the ability to uncover long-term trends and address safety concerns more effectively.

### 3.2 Monitor: Pedestrian Activity Monitoring

Monitoring activity of pedestrians requires two essential blocks, detection and tracking, to be able to track each pedestrian and extract their corresponding trajectory to estimate the speed and heading. Additionally, for calculating surrogate safety measures (SSM) between pedestrians and right-turning vehicles, which is the main cause of pedestrian-vehicles crashes, vehicle detection and tracking is also required. After investigating state-of-the-art deep learning models and computer vision techniques and conducted comparisons among many object detection models, RT-DETR is the best candidate for object detection. RT-DETR is the improved model architecture of DETR and its large model has achieved the highest average precision score (AP) and processing speed on COCO val2017 outperforming the state-of-the-art YOLO detectors and DETRs in both speed and accuracy (Zhao et al., 2024). For tracking, bytetrack is the fittest tracking algorithm for this system, since it has achieved higher performance metric than many other tracking methods on MOT17, MOT20, and Multiple Object Tracking in Highway Videos (Abouelyazid, 2023; Zhang et al., 2022). RT-DETR and ByteTrack will be used to monitor the movements of pedestrians at intersections and right-turning vehicles to extract their corresponding trajectories.

#### 3.2.1 Conflict Analysis using Surrogate Safety Measure

By extracting the trajectories of the pedestrians and right-turning vehicles, Surrogate Safety Measures (SSMs) can be derived, and in this work, we will specifically focus on the Time to Collision (TTC) metric (Lee, 1976). The decision to use only TTC is motivated by the need to maintain the monitoring system and report generation processes in real-time, as calculating multiple SSMs requires significant computational resources. TTC is calculated when there is a potential interaction between pedestrians and right-turning vehicles, estimating the time remaining before a collision would occur if the current trajectories continued unchanged. By analyzing TTC, we can identify and highlight any recurring safety concerns at the intersection involving pedestrians and right-turning vehicles (Hagiwara et al., 2008; Vogel, 2003; Xin et al., 2021). The values of TTC are categorized as follows: less than 1.5 seconds: serious conflict, between 1.5 and 3 seconds: slight conflict, more than 3 seconds no conflict. TTC provides valuable insights into the dynamics of pedestrian-vehicle interactions and helps pinpoint areas where safety improvements may be necessary, thereby enhancing pedestrian safety and mitigating risks at intersections with frequent right-turning vehicle movements.

**3.2.2 Crossing Violation Detection**

One of the critical safety events to capture is pedestrian crossing violations at signalized intersections. These violations expose pedestrians to conflicts with approaching vehicles, potentially leading to serious crashes (Zhang et al., 2021). To determine if a pedestrian crossed during the designated signal phase, pedestrian phase signal timings will be imported via the Automated Traffic Signal Performance Measures (ATSPM) API provided by FDOT services. ATSPM provides the start and end times of the pedestrian signal phase. Fig. 2 illustrates the detection process for crossing violations: if a pedestrian crosses during the pedestrian phase, it is considered a safe crossing. However, if the pedestrian crosses outside the designated phase, it is identified as a crossing violation. This information is essential for analyzing pedestrian safety and implementing measures to reduce potential conflicts and accidents.

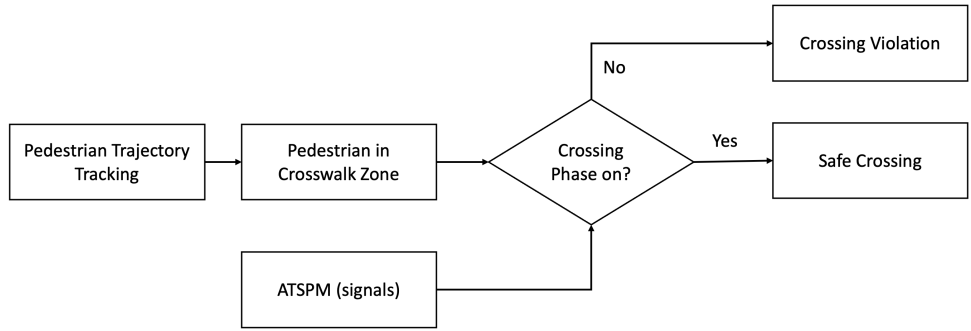

**Fig. 2. Crossing Violation Detection Diagram**

**3.3 Reporter: Pedestrian Activity Report Generation**

To generate real-time reports on pedestrian activity, fast large language models (LLMs) are essential. Although there are many relatively small LLMs available, not all achieve high performance on well-known benchmarks such as the Measuring Massive Multitask Language Understanding (MMLU) dataset (Hendrycks et al., 2020). Several small in size LLMs were selected for testing their inference speed (tokens/second) and reasoning quality (MMLU score). As shown in Fig. 3, the phi3-mini LLM stands out as the most capable among the small LLMs, with a context window of 4k, which is sufficient for generating reports on pedestrian activity. Notably, this powerful yet compact LLM, with its high reasoning score compared to larger models, can be deployed on edge devices, promoting sustainability and enhancing edge computing capabilities.

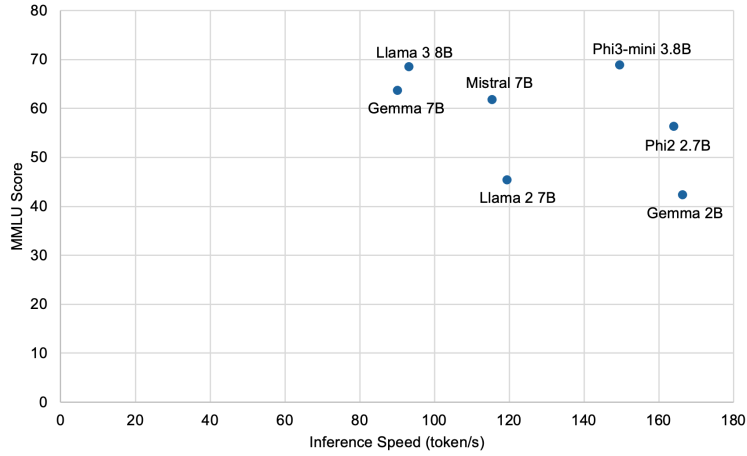

**Fig. 3. Comparison of smaller LLMs based on reasoning score and inference speed**

### 3.4 Analyzer: Historical Analysis

For comprehensive historical analysis of the generated pedestrian activity reports, an LLM with long context window is required with high reasoning capabilities to process large number of reports and extract anomalies, patterns, safety concerns, and other insights. Choosing the best LLM for this task involves comparing various large models based on their performance and capabilities. Fig. 4 presents the conducted comparison between eight LLMs, showing that the Phi-3 medium LLM achieves a high reasoning score on the MMLU benchmark more than most of the other LLMs while offering the longest context window among the models tested. This makes Phi-3 medium particularly suitable for in-depth historical analysis, as it can effectively handle extensive database and provide meaningful insights and patterns.

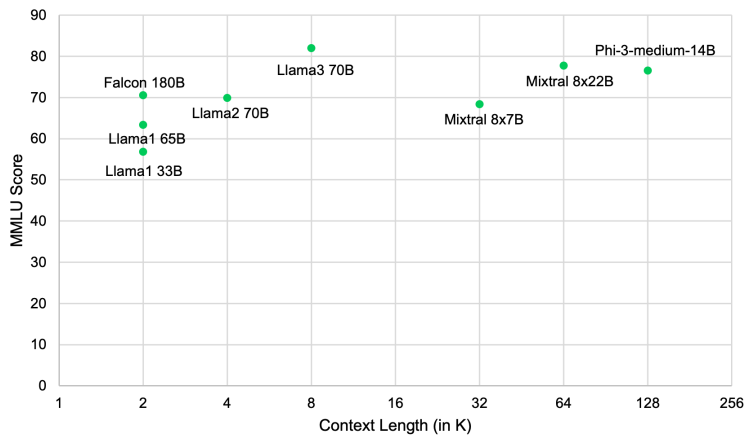

**Fig. 4. Comparison of larger LLMs based on reasoning score and context length**

## 3.5 LLM Fine Tuning

LLMs are trained on general knowledge to excel in many tasks, however, being more domain specific is needed to make the LLM more familiar with traffic safety and pedestrian safety domain. Therefore, a custom dataset has been gathered including information and extracted from conducted studies and analysis on traffic safety at intersections. Due to the large size of Phi-3 mini and Phi-3 medium models, full fine-tuning is computationally intensive. A more efficient method is Adapter-Based Performance Efficient Fine-Tuning (PEFT), specifically the low-rank adaptation (LoRA) technique (Hu et al., 2021). LoRA performs as well as or better than full fine-tuning and prevents overfitting and catastrophic forgetting (Biderman et al., 2024). To further reduce computational demands, quantization is used, converting model parameters from higher bit-width formats to lower ones, like 8-bit integers. This approach, combined with LoRA, is known as QLoRA (Quantized Low-Rank Adapter). QLoRA uses 4-bit NormalFloat (NF4) and double quantization, significantly reducing memory usage without sacrificing performance (Dettmers et al., 2024). **Fig. 5.** illustrates the QLoRA fine-tuning process for the LLMs. The process begins with loading a JSON database of question-and-answer pairs from over 200 studies. The base model is then loaded and the dataset tokenized. The model is quantized from FP32 to NF4, and LoRA parameters are set for fine-tuning. During this process, the base model's parameters remain unchanged, and only the adapters are updated. After finishing the fine-tuning, the LoRA adapters will be merged with the base LLM to be ready for deployment. The fine-tuning process through QLoRA is conducted on an Nvidia A10G 24 GB GPU and took approximately 25 minutes to fine-tune Phi-3-mini and 1.22 hours to finish fine-tuning Phi-3-medium.

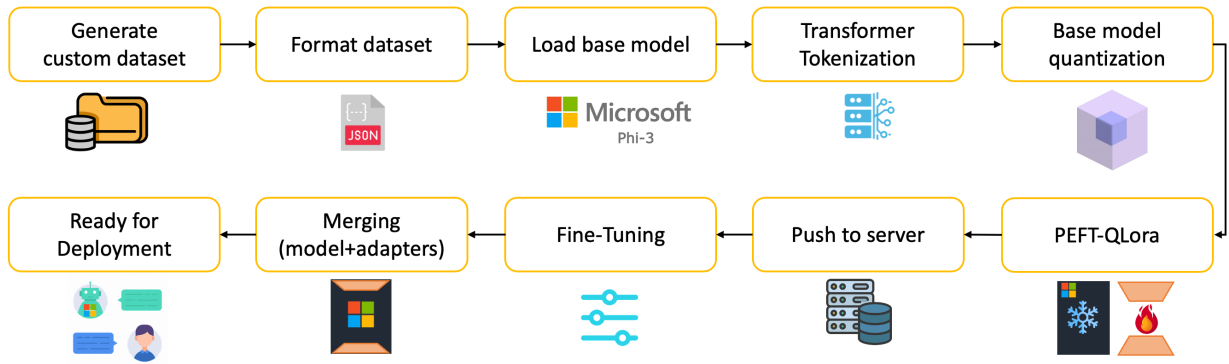

**Fig. 5. Process of fine-tuning Phi-3 mini and medium using QLoRA**

## 4. Results

### 4.1 Experiment Results of the Pedestrian Activity Monitoring and Reporting

This section provides the analysis of pedestrian activity monitoring over a 4-hour video stream (2 hours in daytime and 2 hours in nighttime), capturing the movements of 73 pedestrians, and generating textual reports about their activity every hour as shown in Fig. 6. The analysis revealed numerous instances of crossing violations. While most of these violations did not lead to conflicts, there was one significant incident involving a pedestrian and a right-turning vehicle, where the pedestrian was at fault for crossing before the pedestrian signal was activated. Using the coding capabilities of the Phi-3-medium LLM, it generated the charts shown in Fig. 7 and Fig. 8. These charts analyze the generated reports of the pedestrian activity, highlighting safety-critical events and providing informative visualization of pedestrian activity at intersection. Additionally, they show the percentages of crossing violations versus safe crossings per crosswalk, breaking down the activity by direction. As shown in Fig. 7, crosswalk A experienced higher pedestrian traffic compared to crosswalk B. However, crosswalk B exhibited a significantly higher rate of crossing violations, with 72.2% of cases being violations, compared to only 18.2% at crosswalk A. This high percentage of violations at crosswalk B was primarily due to an hour of rainfall, which influenced pedestrians to violate the signal and cross. This analysis underscores the importance of targeted interventions to improve pedestrian safety, particularly at crosswalk B. Furthermore, most of the crossing violations happened during nighttime with 76% compared to 24% during daytime, and raining influence pedestrians to violate the signal, as visualized in the generated charts in Fig. 8.

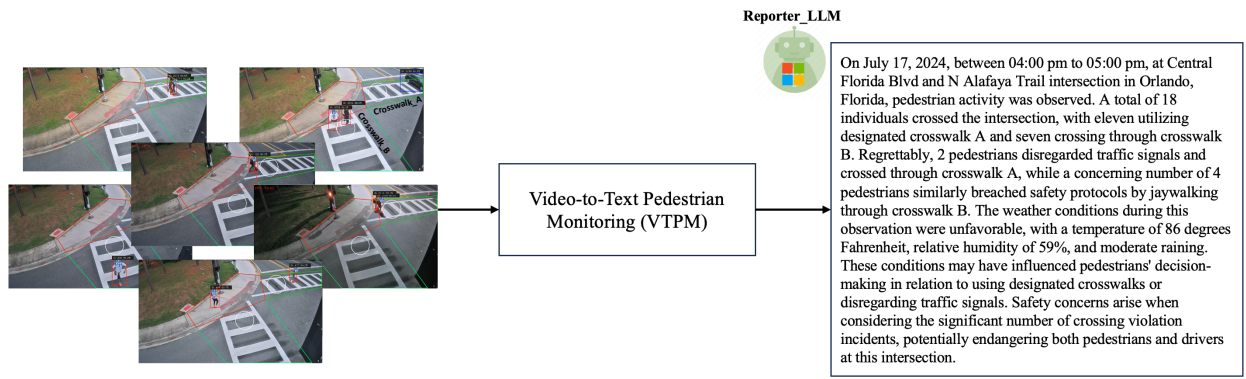

**Fig. 6. Generating textual report about pedestrian activity at intersection for one hour**

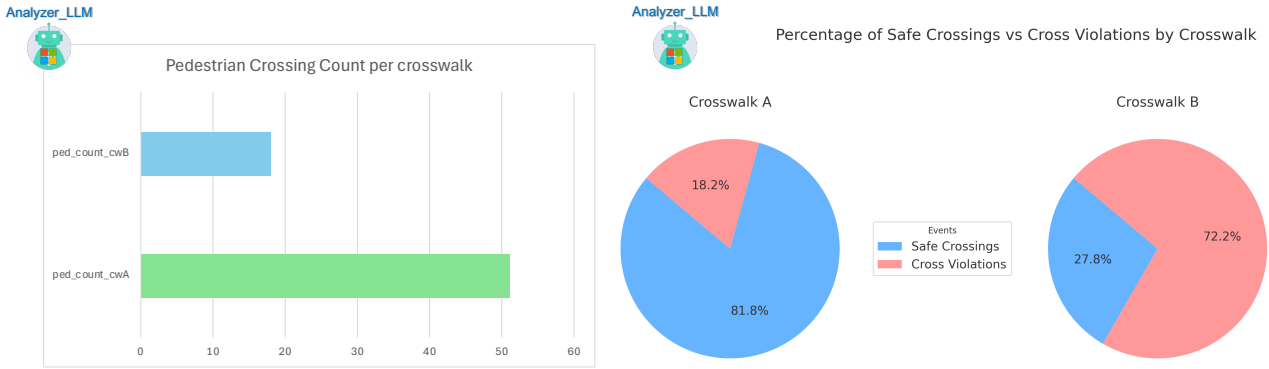

**Fig. 7. LLM generating charts analyzing the crossing violations per crosswalk**

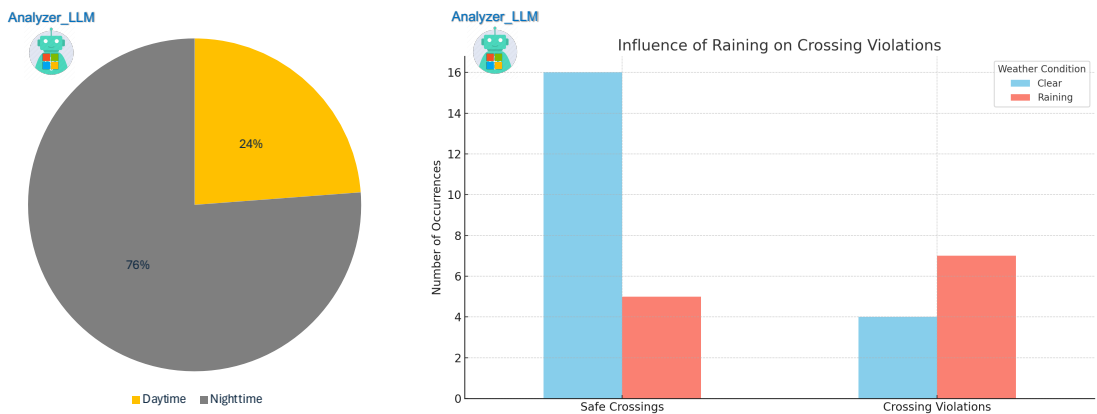

**Fig. 8. LLM generating charts comparing crossing violations between daytime and nighttime and between clear and raining weather conditions**

The computer vision-based monitoring system demonstrated a notable latency of 0.05 seconds per video frame, equating to a processing speed of 20 Frames Per Second (FPS). This performance qualifies the system for real-time video processing, allowing it to promptly analyze and react to pedestrian activities as they unfold. This low latency is critical for ensuring immediate responses and accurate monitoring. Furthermore, the time required to generate reports using the phi-3-mini system was determined to be 0.33 seconds, while generating charts might take about 5-7 seconds for each chart. These reports compile and analyze the monitored data, providing comprehensive insights into pedestrian behavior. The rapid report generation process enables the aggregation of data, yielding meaningful reflections that are invaluable for improving pedestrian safety and informing urban planning strategies. The system demonstrates a high performance in detecting crossing violations, achieving an accuracy of 90.2%, a precision of 91.96%, and a recall of 93.2%. However, some challenges were encountered, such as instances where pedestrians were occluded by vehicles or where pedestrians crossed outside of the designated crosswalk areas. In this experiment, reports are generated every hour, detailing various aspects of pedestrian activity, including pedestrian counts, crossing violations, and conflicts between pedestrians and right-turning vehicles. This hourly reporting ensures a continuous and detailed understanding of pedestrian dynamics. The integration of real-time processing with swift report generation highlights the system's potential for effective and timely monitoring of pedestrian activities in diverse urban environments. This capability is essential for enhancing pedestrian safety and optimizing urban planning efforts, ensuring that interventions can be made quickly and based on accurate, up-to-date information while preserving the privacy of the road users.

## 4.2 Human Evaluation of Analytical Text Generation by LLM

To ensure that the fine-tuned Phi-3 medium (The Analyzer) is capable of analyzing the generated reports by the Reporter as shown in Fig. 6 reflecting the traffic safety concerns of the pedestrian activity, a historical analysis was conducted using data of another day capturing the pedestrian activity over 15 hours (8 am to 11 pm) with varying rain intensities. This analysis aimed to assess the influence of weather on pedestrian behavior and evaluate the system's ability to identify critical safety conditions. The goal was to verify the reasoning and summarization capabilities of the Analyzer in comparison to the base model and other high-performing LLMs. From the previously shown Fig. 4, Llama3 70B achieved the highest MMLU score, followed by Mixtral 8x22B and the pretrained Phi-3 medium 14B model. For qualitative evaluation, 15 reports were generated, summarizing pedestrian activity every hour, including crossing events, violations, conflicts with right-turning vehicles, and weather conditions. A questionnaire was conducted with 16 doctoral and post-doctoral researchers in the field of traffic safety using three evaluation metrics to verify the quality of the fine-tuned phi-3 medium (Analyzer) for historical analysis over several reports covering almost a whole day. The first metric assessed the clarity and accuracy of the textual reports, which were formatted as,

[date, time, and location] [weather condition] [pedestrians count] [crossing violations] [conflicts] to ensure consistency among the reports. These reports summarized 15 hours of pedestrian activity from 8 am to 11 pm. The second metric evaluated the reasoning capability of the generated reports, focusing on the logical coherence and inference quality. The third metric confirmed that the fine-tuned LLM had superior knowledge of pedestrian safety compared to pretrained models, particularly in response to domain-specific questions derived from a custom dataset. Fig. 9 displays the generated textual analysis of these 15 reports, highlighting the safety-critical events associated with pedestrian activity at the intersection and how rain intensity influenced pedestrians to multiple crossing violations. Table 1 shows the numerical results of the questionnaire, revealing that the researchers preferred the textual analytical report generated by the fine-tuned Phi-3-medium (Analyzer) over the other LLMs in terms of analytical summarization and traffic safety knowledge. However, Llama3 70B achieved the highest reasoning score, aligning with its highest reasoning score on the MMLU dataset.

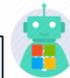

**Fig. 9. Textual analytical summarization of the 15 reports using the analyzer component highlighting the safety critical events**

Table 1. Human evaluation scores among the four LLMs used in the questionnaire

| Evaluation Metric \ LLM | Fine-Tuned Phi3-medium 14B (Ours) | Phi3-medium 14B (Pretrained) | Mixtral 8x22B | Llama3 70B |
|---|---|---|---|---|
| Analytical Summarization | **39%** | 26% | 30% | 4% |
| Reasoning | 21% | 11% | 32% | **37%** |
| Traffic Safety Knowledge | **58%** | 5% | 11% | 26% |

## 5. Conclusions and Future Recommendations

VTPM for pedestrian activity monitoring and reporting system demonstrates effectiveness for privacy-preserve monitoring framework by converting the processed videos using computer vision techniques to textual report using LLMs. With a latency of 0.05 seconds for processing each video frame, the system ensures prompt and real-time analysis. The rapid report generation time of 0.33 seconds enhances effective monitoring and immediate response. The analysis of a 4-hour video stream revealed significant pedestrian activity and numerous crossing violations, especially at night (76%). Crosswalk B exhibited a higher rate of violations (72.2%) compared to crosswalk A (18.2%), indicating the need for targeted safety interventions. A historical analysis validated the improved performance of the fine-tuned LLM (Phi-3 medium) against other models, with Llama3 70B achieving the highest MMLU score. Qualitative evaluations by 16 doctoral transportation researchers showed the fine-tuned LLM excelled in clarity, accuracy, reasoning capability, and domain knowledge. Overall, the system's combination of real-time processing and swift report generation underscores its effectiveness for timely monitoring of pedestrian activities. This capability is essential for improving pedestrian safety and optimizing urban planning, allowing for prompt interventions based on accurate data while preserving pedestrian privacy. Moreover, by converting video footage into textual reports, the system dramatically reduces memory usage by up to 253 million percent, resolves privacy concerns, and facilitates comprehensive, interactive historical analysis for transportation departments and research organizations. For future work, it is recommended to integrate VTPM with cameras at all corners of the intersection. This would allow for comprehensive coverage of pedestrian activity across the entire intersection, enabling the capture of more events and providing a more detailed analysis of pedestrian behavior and safety.

**Acknowledgment**

The authors acknowledge funding and support from the Florida Department of Transportation (FDOT).